\documentclass[times,review,10pt]{elsarticle}

\usepackage{amssymb}
\usepackage{amsmath}

\usepackage{times}
\usepackage{epsfig}
\usepackage{graphicx}
\usepackage{bm}
\usepackage{tabu}
\usepackage{verbatim}
\usepackage{multirow}
\usepackage{gensymb}
\usepackage{color}
\usepackage{marvosym}
\usepackage{colortbl}
\usepackage{makecell}
\usepackage[normalem]{ulem}
\usepackage{balance}
\usepackage{arydshln}

\usepackage{anyfontsize}

\definecolor{color_1st}{rgb}{0.39, 0.75, 0.48}
\definecolor{color_2nd}{rgb}{0.57, 0.80, 0.52}
\definecolor{color_3rd}{rgb}{0.75, 0.86, 0.56}
\definecolor{color_4th}{rgb}{0.93, 0.92, 0.60}
\definecolor{color_5th}{rgb}{1.00, 1.00, 1.00}
\definecolor{color_6th}{rgb}{1.00, 1.00, 1.00}
\definecolor{color_7th}{rgb}{1.00, 1.00, 1.00}

\journal{Elsevier Journal}

\begin{document}

\begin{frontmatter}



\title{Incremental Rotation Averaging Revisited}


\author{Xiang Gao$^{a,b}$, Hainan Cui$^{b}$, Yangdong Liu$^{a,b}$, and Shuhan Shen$^{a,b,c,*}$} 

\affiliation{organization={Luoyang Institute for Robot and Intelligent Equipment},
            city={Luoyang},
            postcode={471000}, 
            country={China}}
\affiliation{organization={Institute of Automation, Chinese Academy of Sciences},
            city={Beijing},
            postcode={100190}, 
            country={China}}
\affiliation{organization={School of Artificial Intelligence, University of Chinese Academy of Sciences},
            city={Beijing},
            postcode={100049}, 
            country={China}}

\begin{abstract}
In order to further advance the accuracy and robustness of the incremental parameter estimation-based rotation averaging methods, in this paper, a new member of the Incremental Rotation Averaging~(IRA) family is introduced, which is termed as IRAv4. As its most significant feature, a task-specific connected dominating set is extracted in IRAv4 to serve as a more reliable and accurate reference for rotation local-to-global alignment. This alignment reference is incrementally constructed, together with the absolute rotations of the vertices belong to it simultaneously estimated. Comprehensive evaluations are performed on the 1DSfM dataset, by which the effectiveness of both the reference construction method and the entire rotation averaging pipeline proposed in this paper is demonstrated.
\end{abstract}




\end{frontmatter}



\section{Introduction} \label{sec:1}

Image-based large-scale scene 3D reconstruction is a fundamental task in computer vision community, which has been widely investigated in recent years~\cite{Gao-ISPRS-18, Gao-TCSVT-20, Gao-TCSVT-22-2, Gao-TITS-23}. As its core step, Structure from Motion~(SfM)~\cite{Cui-ICCV-15, Schonberger-CVPR-16, Cui-CVPR-17, Cui-3DV-17, Zhu-CVPR-18, Manam-ECCV-22, Cui-TIP-23} aims to simultaneously recover camera pose and scene structure given pair-wise image feature matches. According to the camera pose initialization scheme, SfM methods could be roughly divided into incremental and global ones. The camera poses are sequentially estimated in incremental SfM via a iterative optimization pipeline~\cite{Schonberger-CVPR-16, Cui-3DV-17}, but are simultaneously solved in global SfM by the motion averaging methodology~\cite{Cui-ICCV-15, Zhu-CVPR-18}. Compared with incremental SfM methods, global ones feature fewer optimization iterations and estimation parameters, which makes them with more advantage in theory and greater potential for application when dealing with increasingly larger reconstruction scene scales.

Motion averaging, which takes relative camera motions (rotations and translations) as input and produces absolute camera poses (orientations and locations), is the primary technique used for camera pose recovery in the global SfM. Due to scale ambiguity of the relative translation estimated via essential matrix decomposition~\cite{Nister-TPAMI-04}, motion averaging is conducted in the manner of first rotation and then translation averaging in most cases~\cite{Govindu-CVPR-01, Martinec-CVPR-07, Crandall-TPAMI-13, Moulon-ICCV-13, Zhu-CVPR-18, Pan-ECCV-24-1}. As the former of the above two phases, rotation averaging~\cite{Hartley-IJCV-13} directly influences the effect of the subsequent translation averaging phase~\cite{Ozyesil-CVPR-15}, and even all the remaining 3D reconstruction procedure, \emph{e.g.}~multi-view triangulation for scene recovery and global Bundle Adjustment~(BA) for parameter optimization. Though increasing attention has been drawn in recent years~\cite{Wilson-CVPR-20, Dellaert-ECCV-20, Eriksson-TPAMI-21, Chen-CVPR-21, Parra-CVPR-21, Moreira-ICCV-21, Zhang-CVPR-23, Li-CVPR-24, Pan-ECCV-24-2}, the problem of rotation averaging is still far from being solved due to the issues of large scale, imbalanced connectivity, and high-level noise of the Epipolar-geometry Graph~(EG).

Recently, a series of rotation averaging methods~\cite{Gao-IJCV-21, Gao-TCSVT-22-1, Gao-TCSVT-23, Gao-TCSVT-24} are proposed based on the ideology of incremental parameter estimation stemmed from the incremental SfM. As the primitive and primary method, Incremental Rotation Averaging~(IRA)~\cite{Gao-IJCV-21} performs incremental absolute rotation computation and relative rotation outlier filtering simultaneously, by which the rotation estimation accuracy and robustness are both guaranteed. To further enhance the efficiency and scalablility, IRA++~\cite{Gao-TCSVT-22-1} is proposed, where the input EG is clustered to construct several low-level intra-sub-EGs and a high-level inter-one. Then IRA is performed on all the (intra- and inter-) sub-EGs to achieve local estimation and global alignment of absolute rotations, respectively. In order to achieve a task-specific EG clustering for better rotation averaging performance, IRAv3~\cite{Gao-TCSVT-23} is presented, where the cluster affiliation of each camera is dynamically determined with its absolute rotation (in the local coordinate system of intra-sub-EG) simultaneously estimated. To accomplish a better global alignment of the local rotation estimates, inspired by the method of WCDS~\cite{Jiang-TGRS-22} proposed by Jiang~\emph{et~al.}, IRAv3+~\cite{Gao-TCSVT-24} is introduced. Instead of performing a cluster-level rotation averaging, which is done in both IRA++ and IRAv3, multiple Connected Dominating Sets~(CDSs) are randomly extracted to serve as the reference for rotation local-to-global alignment. 

However, in this paper we further argue that the accuracy and robustness of the IRA series described above could be further advanced. Though the effectiveness of the cluster-based pipeline with task-independent CDS serving as global reference has been demonstrated in both IRAv3+~\cite{Gao-TCSVT-24} and WCDS~\cite{Jiang-TGRS-22}, we believe that with a task-specific CDS extracted, more reliable global reference construction together with more accurate camera pose globalization would be further achieved. Based on the above analysis, a novel rotation averaging method termed as IRAv4 is proposed in this paper, which is built upon IRAv3+. The major difference between them lies in that instead of first extracting multiple CDSs and then estimating the absolute rotations in the CDSs' local coordinate system by leveraging IRA, which is done in IRAv3+, the (task-specific) CDS is extracted by incrementally selecting the Next-Best Vertex~(NBV) to maximize the supports from the currently extracted ones in the CDS, together with its absolute rotation (in the local coordinate system of CDS) simultaneously estimated. Our IRAv4 is evaluated on the 1DSfM dataset and compared with several other mainstream rotation averaging methods to demonstrate its effectiveness and advantages.

\section{Related works}

{Except for the IRA series, there are mainly three kinds of large-scale rotation averaging methods, including the robust loss-based~\cite{Chatterjee-TPAMI-18, Shi-ICML-20, Shi-ICML-22}, the outlier filtering-based~\cite{Cui-ISPRS-19, Gao-SPL-20, Lee-CVPR-22}, and the deep learning-based~\cite{Purkait-ECCV-20, Yang-CVPR-21, Li-CVPR-22} ones, which are briefly reviewed in the following.}

\subsection{Robust loss-based rotation averaging}

{The robust loss-based methods~\cite{Chatterjee-TPAMI-18, Shi-ICML-20, Shi-ICML-22} seek to design robust loss functions to achieve a robust estimation of the absolute camera rotations in the presence of relative rotation measurement outliers. Chatterjee~\emph{et~al.}~\cite{Chatterjee-TPAMI-18} developed a two-stage rotation averaging method. They firstly use the $\ell_1$ solution as an initialization to make the method robust to outliers. Then, an Iteratively Reweighted Least Squares (IRLS) approach based on the $\ell_\frac{1}{2}$ loss function is followed to achieve an efficient and accurate absolute rotation estimation. Shi~\emph{et~al.}~\cite{Shi-ICML-20} proposed a novel robust optimization framework, Message Passing Least Squares (MPLS), as an alternative paradigm to IRLS for rotation averaging, and the advantages of MPLS over IRLS are theoretically analyzed. After that, based on the cyclic consistency constraint, Shi~\emph{et~al.}~\cite{Shi-ICML-22} presented a novel quadratic programming framework, Detection and Estimation of Structural Consistency~(DESC), which is less dependent to initialization and more simple in optimization. Though the above methods are robust to the relative rotation outliers to some extent, the optimization procedures in them still heavily depend on the initialization and are usually inefficient because of the relatively complicated problem formulations.}

\subsection{Outlier filtering-based rotation averaging}

{Compared with the robust loss-based methods, the outlier filtering-based ones~\cite{Cui-ISPRS-19, Gao-SPL-20, Lee-CVPR-22}, are more theoretically straightforward. Cui~\emph{et~al.}~\cite{Cui-ISPRS-19} proposed a multiple Orthogonal Maximum Spanning Trees~(OMSTs)-based rotation averaging method. Multiple OMSTs are selected from the input EG to obtain a set of densely connected and more accurate relative rotation measurements for robust rotation averaging~\cite{Chatterjee-TPAMI-18}. Gao~\emph{et~al.}~\cite{Gao-SPL-20} presented a Hierarchical RANSAC-based Rotation Averaging (HRRA) method. The absolute rotations are estimated based on the thought of RANSAC with random spanning tree of the input EG serving as the minimal set for model estimation. In addition, graph clustering is employed to downsize the spanning trees as the RANSAC-based parameter estimation is sensitive to the size of the minimal set. Lee~\emph{et~al.}~\cite{Lee-CVPR-22} introduced the algorithm of HARA, which incrementally adds constraints based on the size of the supporting camera triplets to construct EG's spanning tree and remove outliers in relative rotation measurements, thus achieving a hierarchical and robust initialization of camera absolute rotation. Though effective outlier filtering and accurate rotation estimation could be achieved in most cases, we find that the performance of the above methods exhibit some instability when dealing with the EGs with relatively high noise level.}

\subsection{Deep learning-based rotation averaging}

{Recently, several deep learning-based methods have been proposed dealing with the rotation averaging problem by leveraging the techniques of graph neural networks~\cite{Purkait-ECCV-20, Yang-CVPR-21, Li-CVPR-22}. Purkait~\emph{et~al.}~\cite{Purkait-ECCV-20} proposed NeuRoRA, a two-stage graph neural networks for rotation averaging. The first stage is a view-graph cleaning network to detect relative rotation measurement outliers. The second stage is a fine-tuning network to refine the absolute orientations. Yang~\emph{et~al.}~\cite{Yang-CVPR-21} presented Multi-Source Propagator~(MSP), which takes appearance information as input and introduces a differentiable Shortest Path Tree~(SPT) method to achieve a robust initialization result. The initialization is further improved through non-learnable iterative edge reweighting. Li~\emph{et~al.}~\cite{Li-CVPR-22} proposed the method of Rotation Averaging Graph Optimizer~(RAGO), a learnable iterative graph optimizer minimizing a gauge-invariant cost function with an edge rectification strategy to mitigate the effect of inaccurate measurements. However, these deep learning-based methods depend heavily on the training data, which makes their generalization and scalability poor in novel situations.}

\section{Brief description on the previous IRA series} \label{sec:2}

Before introducing the proposed IRAv4 method in this paper, brief descriptions on the existing IRA series, including IRA~\cite{Gao-IJCV-21}, IRA++~\cite{Gao-TCSVT-22-1}, IRAv3~\cite{Gao-TCSVT-23}, and IRAv3+~\cite{Gao-TCSVT-24}, are provided for their better understanding. And more details on them could be found in the original papers.

\noindent\textbf{IRA}~\cite{Gao-IJCV-21} mainly has two steps: 1)~The camera triplet with minimum cyclic rotation deviation after local optimization is selected as the initial seed and its optimized absolute rotations (in the camera triplet's local coordinate system) are served as the seed estimation. 2)~The camera with most supporting EG edges during chaining-based absolute rotation pre-computation is selected as the NBV and the pre-computed absolute rotation is served as its initialization; and then, either local (on the newest estimated absolute rotation only) or global optimization (on all the currently estimated absolute rotations) is performed. Note that inlier/outlier relative rotation measurements related to the absolute rotations to be optimized could be distinguished based on their current estimates, and only the inliers are involved in the above optimization. The NBV selection, initialization, and optimization are iteratively performed until all the absolute rotations have been estimated.

\noindent\textbf{IRA++}~\cite{Gao-TCSVT-22-1} contains five steps: 1)~Community detection-based EG clustering~\cite{Zhou-ECCV-20} is carried out on the input EG to obtain several intra-sub-EGs. 2)~IRA is performed on each low-level intra-sub-EG to estimate the cameras' absolute rotations in its local coordinate system. 3)~Voting-based single rotation averaging~\cite{Gao-SPL-20} is conducted to estimate the relative rotation between the local coordinate systems of each intra-sub-EG pair. 4)~IRA is performed again on the high-level inter-sub-EG to estimate the absolute rotation of each intra-sub-EG's local coordinate system. 5)~And finally, rotation global alignment and optimization is conducted to first globally align the absolute rotations of all the vertices in the input EG to a uniform coordinate system, and then globally optimize them to produce the final rotation averaging result.

\noindent\textbf{IRAv3}~\cite{Gao-TCSVT-23} comprises five steps as well, with its last three steps similar to those of IRA++. And for the first two: 1)~Community detection-based seed construction is performed to construct several cluster seeds for the follow-up on-the-fly procedures in its second step. 2)~On-the-fly EG clustering and intra-sub-EG rotation estimation are conducted to dynamically assign unregistered vertices to certain EG clusters and iteratively estimate their absolute rotations in their assigned clusters' local coordinate systems. The second step is the core one of IRAv3 and contains three sub-steps, including potential cluster and vertex pre-selection, NBV selection and cluster affiliation, and incremental absolute rotation computation, with the last two iteratively performed for the on-the-fly procedures.

\noindent\textbf{IRAv3+}~\cite{Gao-TCSVT-24} consists of five steps once again, which shares the same first two and last one steps with IRAv3 for 1)~and 2)~on-the-fly EG clustering and intra-sub-EG rotation estimation, and 5)~rotation global alignment and optimization, respectively. After executing the first two steps for dynamic vertex cluster affiliation and rotation estimation, the third and forth ones are sequentially conducted: 3)~Multiple CDSs are randomly extracted and IRA is performed on the CDSs-based sub-EG for global alignment reference construction. 4)~Cluster-wise relative rotation for rotation local-to-global alignment is estimated by leveraging the guidance of the cluster-to-reference common vertices. Subsequently, similar to both IRA++ and IRAv3, the local absolute rotations in each cluster are firstly globally aligned to a uniform coordinate system (that of the CDSs-based sub-EG for IRAv3+), and then globally optimized.

\section{The proposed IRAv4 method} \label{sec:3}

The proposed IRAv4 in this paper is directly built upon the IRAv3+~\cite{Gao-TCSVT-24}, and the major difference between them is the manner of global reference construction. It should be noted that the process of global reference construction involves two sub-steps of 1)~reference-based sub-EG extraction and 2)~sub-EG rotation estimation. Based on the description in the last section, the above two sub-steps in IRAv3+ are conducted sequentially to result in a task-independent global reference. However, for our proposed IRAv4, it follows the incremental parameter estimation pipeline during the global reference construction, where the vertices in the reference-based sub-EG are incrementally selected together with their absolute rotations simultaneously estimated. By this way, a task-specific reference for rotation local-to-global alignment is constructed. The process of the task-specific reference construction is detailed in the following. Before that, the rotation averaging problem is briefly formulated for better understanding at first: The input EG is denoted as $\mathcal{G}=(\mathcal{V},\mathcal{E})$, where $\mathcal{V}$ and $\mathcal{E}$ denote the sets of cameras and camera pairs with sufficient image local feature match inliers for essential matrix estimation, and the rotation averaging problem is defined as given the relative rotation measurements $\{\boldsymbol{R}_{i,j}|e_{i,j}\in\mathcal{E}\}$, and estimating the absolute camera orientations $\{\boldsymbol{R}_{i}|v_{i}\in\mathcal{V}\}$.

\begin{figure}
\centering
\includegraphics[width=0.73\columnwidth]{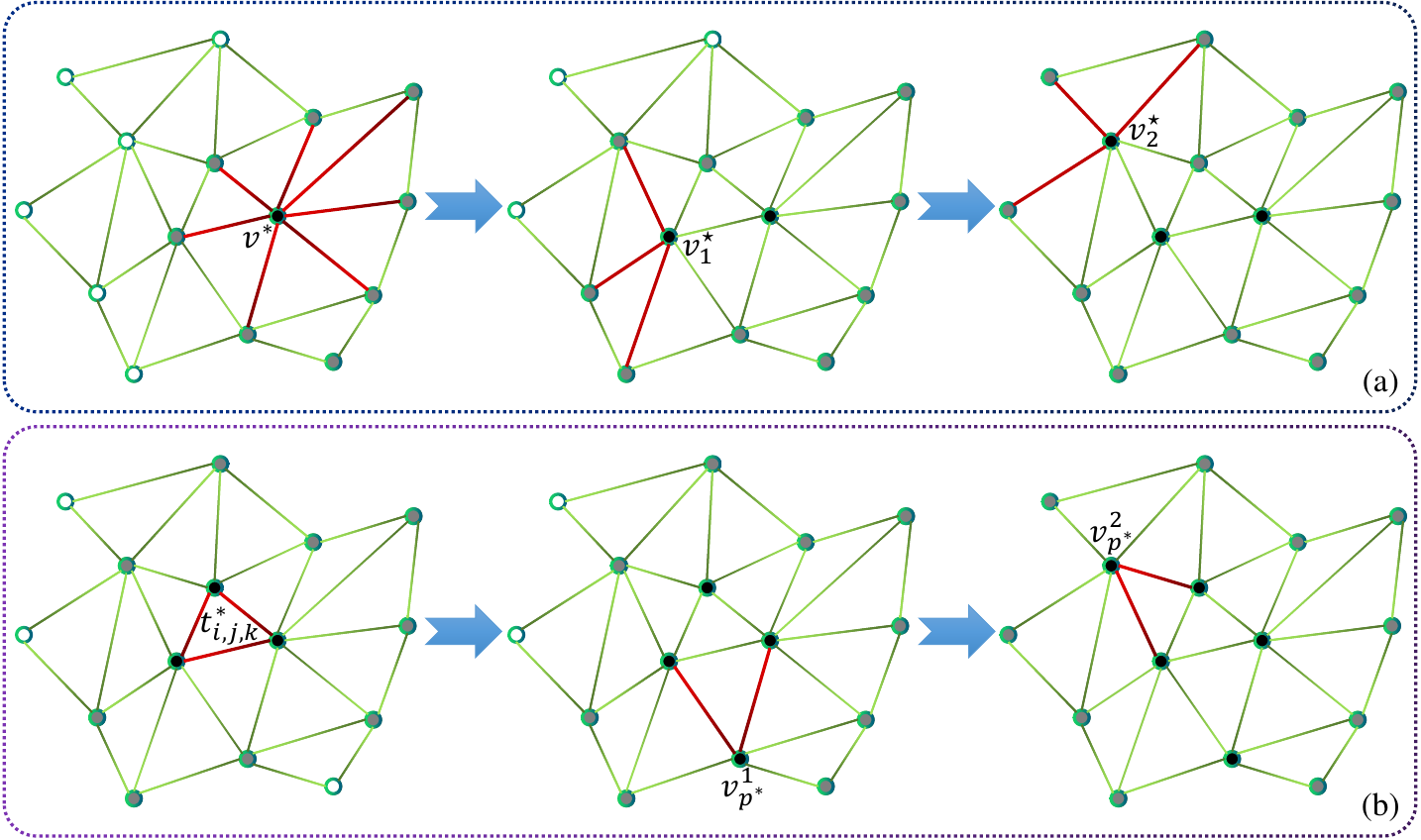}
\caption{Toy examples of the traditional method (a) based on an approximation algorithm~\cite{Guha-Algorithmica-98} and the task-specific method (b) proposed in IRAv4 for Connected Dominating Set~(CDS) extraction. For (a), the red edges denote those between the currently selected vertex ($v^*$ or $v_i^\star$, where $v_i^\star$ denotes the selected vertex in the $i$-th iteration) and its adjacent unselected ones. And for (b), the red edges denote those between the selected triplet $t_{i,j,k}^*$ and the edge supporting set for the selected Next-Best Vertex~(NBV) $v_{p^*}^i$ in the initialization step and the $i$-th iteration step, respectively. It could be observed from the figure that the number of vertices in the task-specific CDS is usually larger than that of the CDS extracted in the traditional way (5~\emph{vs.}~3 in this toy example). Please refer to the main text for more details.}
\label{fig:1}
\end{figure}

\subsection{Task-specific alignment reference construction}

Similar to WCDS~\cite{Jiang-TGRS-22} and IRAv3+~\cite{Gao-TCSVT-24}, CDS-based sub-EG is also used to serve as a global reference for camera pose alignment in IRAv4. The CDS problem in graph theory asks for a \emph{minimum-size} and \emph{connected} subset of vertices with the following property which accounts for the concept of \emph{dominating}: Each vertex is required to either be in the CDS, or adjacent to some vertex of the CDS. The CDS problem is usually solved based on an approximation algorithm~\cite{Guha-Algorithmica-98}, which roughly proceeds as follows:
\begin{enumerate}
    \item[\textbf{Init:}] Mark all vertices in $\mathcal{V}$ white. Select the vertex $v^\ast$ with most neighbours in a weighted or unweighted manner. Mark $v^\ast$ and its neighbours black and gray, respectively.
    \item[\textbf{Iter:}] For all the gray vertices, select the vertex $v^\star$ with most white neighbours in a weighted or unweighted manner. Mark $v^\star$ and its white neighbours in black and gray, respectively. Iterate the above step until there is no white vertex in $\mathcal{V}$, and the black vertex set constitutes the CDS.
\end{enumerate}
Fig.~\ref{fig:1}(a) gives a toy example on the traditional (or task-independent) CDS extraction procedure described above.

However, in this paper we claim that compared with the task-independent CDS extracted based on graph theory, a task-specific CDS would provide a more reliable global reference for camera rotation alignment, by which more accurate and robust rotation estimation could be achieved. The procedure of the task-specific CDS extraction in this paper is stemmed from the pipeline of IRA~\cite{Gao-IJCV-21}, and the major difference between them are the termination conditions. As described in the last section, IRA's iteration of NBV selection, initialization, and optimization stops until all the absolute rotations of the vertices in $\mathcal{V}$ have been estimated and optimized. Nevertheless, the task-specific CDS extraction in IRAv4 shares the termination condition of the traditional CDS extraction to guarantee that it is extracted to exhibit the properties of \emph{connected} and \emph{dominating}, \emph{i.e.}~the vertices in the CDS are connected and all the vertices in $\mathcal{V}$ is either in or adjacent to some vertex of the CDS. Note that during the extraction of the task-specific CDS in IRAv4, the absolute rotations of the extracted vertices are simultaneously estimated, and the property of \emph{minimum} in size is no longer guaranteed. In fact, as the vertex with higher reliability instead of larger coverage is selected with priority, the number of vertices in the task-specific CDS of IRAv4 is always larger than that of the task-independent one extracted by the traditional method~\cite{Guha-Algorithmica-98}. The task-specific CDS extraction procedure for global reference construction, which is illustrated by a toy example in Fig.~\ref{fig:1}(b)), is detailed in the following.

For the \emph{initialization} step, the camera triplet set in $\mathcal{G}$ is obtained and denoted as $\mathcal{T}$, and for each triplet $t_{i,j,k}\in\mathcal{T}$, similar to IRA, the absolute rotations of $\{v_i,v_j,v_k\}$ in the triplet's local coordinate system are initialized and optimized by measurement chaining and residual minimization of the relative rotations, by which the CDS's initial triplet is selected together with the absolute rotations estimated. Specifically, the absolute rotations in $t_{i,j,k}$ are firstly initialized as:
\begin{equation} \label{eq:1}
    \bm{R}_i=\bm{I}, \bm{R}_j=\bm{R}_{i,j}, \bm{R}_k=\bm{R}_{i,k},
\end{equation}
and then fed into the following triplet-based chaining check:
\begin{equation} \label{eq:2}
    d_{\bm{R}}(\bm{R}_{j,k},\bm{R}_{i,k}\bm{R}_{i,j}^{\top})<\theta_{\mathrm{th}},
\end{equation}
where $d_{\bm{R}}(\bm{R}_1,\bm{R}_2)=\arccos\frac{\mathrm{tr}(\bm{R}_2\bm{R}_1^\top)-1}{2}$ is the angular distance between two rotation matrices, $\bm{R}_1$ and $\bm{R}_2$, and $\theta_{\mathrm{th}}=3\degree$ is the outlier threshold in this paper. The triplet set that passes the above chaining check, $\mathcal{T}'$, is involved in the following optimization for initial seed selection and estimation:
\begin{equation} \label{eq:3}
    \bm{R}_i^*,\bm{R}_j^*,\bm{R}_k^*=\arg\min\sum_{\substack{v_i,v_j\in\mathcal{V}_{t_{i,j,k}}\\e_{i,j}\in\mathcal{E}_{t_{i,j,k}}}}d_{\bm{R}}^2(\bm{R}_{i,j},\bm{R}_{j}\bm{R}_{i}^{\top}),
\end{equation}
the above optimization problem and the other ones in the rest of this paper are solved by the Ceres Solver library, and the triplet $t_{i,j,k}^*=\{v_{i^*},v_{j^*},v_{k^*}\}$ with the largest selection reward defined in the following, together with the optimized rotations $\{\bm{R}_{i^*}^*,\bm{R}_{j^*}^*,\bm{R}_{k^*}^*\}$ are served as the initial seed construction:
\begin{equation} \label{eq:4}
    t_{i,j,k}^*=\arg\max_{t_{i,j,k}\in\mathcal{T}'}\sum_{\substack{v_i,v_j\in\mathcal{V}_{t_{i,j,k}}\\e_{i,j}\in\mathcal{E}_{t_{i,j,k}}}}\cos(d_{\bm{R}}(\bm{R}_{i,j},\bm{R}_j^*{\bm{R}_i^*}^\top)).
\end{equation}

For the \emph{iteration} step, the vertex sets of the currently selected and not selected to the global reference are denoted as $\mathcal{V}^s$ and $\mathcal{V}^t$, respectively. Note that the absolute rotations of all the vertices in $\mathcal{V}^s$ have been estimated in $\mathcal{V}^s$'s local coordinate system, which are denoted as $\left\{\bm{R}_m\middle|v_m\in\mathcal{V}^s\right\}$. And the vertex $v_{p^*}$ in $\mathcal{V}^t$ that receives largest supports from $\mathcal{V}^s$ is selected from $\mathcal{V}^t$ to $\mathcal{V}^s$ together with its rotation firstly initialized and then optimized. This iteration step of selection, initialization, and optimization stops until the (connecting and dominating) termination conditions of the task-specific CDS construction reaches, and then the current $\mathcal{V}^s$ together with the estimated absolute rotations of the vertices in $\mathcal{V}^s$ are served as the global reference constructed by IRAv4. Specifically, for each vertex $v_p$ in $\mathcal{V}^t$, the edge set between it and $\mathcal{V}^s$ is obtained and denoted as $\mathcal{E}_p$, and for each edge $e_{m,p}$ in $\mathcal{E}_p$ connecting $v_m$ in $\mathcal{V}^s$ and $v_p$ in $\mathcal{V}^t$, the pre-computing set of $v_p$'s absolute rotation is obtained by:
\begin{equation} \label{eq:5}
    \left\{\bm{R}_p^m=\bm{R}_{m,p}\bm{R}_m\middle|v_m\in\mathcal{V}^s,v_p\in\mathcal{V}^t,e_{m,p}\in\mathcal{E}_p\right\}.
\end{equation}
Though the items in $\{\bm{R}_p^m\}$ represent the absolute rotation pre-computations of the same vertex $v_p$, the ideal identity is hard to hold practically due to the inevitable errors in both $\bm{R}_{m,p}$ and $\bm{R}_m$. To deal with this, similar to IRA, the supporting set of each item in the pre-computing set defined in Eq.~\ref{eq:5} for each vertex $v_p$ in $\mathcal{V}^t$ is leveraged for the iteration step of the task-specific CDS extraction. Specifically, by leveraging the pre-computation of $\bm{R}_p^m$ and the rotation estimations in $\mathcal{V}^s$ connected by $\mathcal{E}_p$, the relative rotations on $\mathcal{E}_p$ could be re-computed and compared with the corresponding measurements for edge supporting set $\mathcal{E}_p^m$ acquisition:
\begin{equation} \label{eq:6}
    \mathcal{E}_p^m=\left\{d_{\bm{R}}(\bm{R}_{n,p},\bm{R}_p^m\bm{R}_n^{\top})<\theta_{\mathrm{th}}\middle|v_n\in\mathcal{V}^s,e_{n,p}\in\mathcal{E}_p\right\}.
\end{equation}
Then, the selection reward for edge $e_{m,p}$ is computed by:
\begin{equation} \label{eq:7}
    \mathrm{reward}(m,p)=\sum_{\substack{v_n\in\mathcal{V}^s\\e_{n,p}\in\mathcal{E}_p^m}}\cos\left(d_{\bm{R}}(\bm{R}_{n,p},\bm{R}_p^m\bm{R}_n^{\top})\right),
\end{equation}
and the edges(vertices) $e_{m^*,p}$($v_{m^*}$) and $e_{m^*,p^*}$($v_{p^*}$) for the pre-computation of vertex $v_p$'s rotation and the NBV determination of iteration step are selected by:
\begin{equation} \label{eq:8}
    \begin{cases}
        v_{m^*}=\arg\max_{v_m\in\mathcal{V}^s}\mathrm{reward}(m,p),\\
        v_{p^*}=\arg\max_{v_p\in\mathcal{V}^t}\mathrm{reward}(m^*,p).
    \end{cases}
\end{equation}
Then, the absolute rotation of the selected NBV $v_{p^*}$ could be initialized by $\bm{R}_{m^*,p^*}\bm{R}_{m^*}$. After NBV selection and initialization, optimization is further performed for accuracy and robustness improvement of the task-specific CDS construction. During the optimization procedure, local optimization on each newly selected NBV $v_{p^*}$ is continuously performed and global optimization on all the currently estimated vertices $\{v_{p^*}\}\cup\mathcal{V}^s$ is intermittently carried out once the size of $\left|\{v_{p^*}\}\cup\mathcal{V}^s\right|$ having grown by a certain rate, say $5\%$. Specifically, for the local optimization, the inlier edge set for providing optimization constraints is first obtained by:
\begin{equation} \label{eq:9}
    \mathcal{E}_{p^*}^{m^*}=\left\{d_{\bm{R}}(\bm{R}_{n,p^*},\bm{R}_{p^*}^{m^*}\bm{R}_n^{\top})<\theta_{\mathrm{th}}\middle|v_n\in\mathcal{V}^s,e_{n,p^*}\in\mathcal{E}_{p^*}\right\},
\end{equation}
where $\bm{R}_{p^*}^{m^*}=\bm{R}_{m^*,p^*}\bm{R}_{m^*}$ is the initialization of the vertex $v_{p^*}$. Then, the absolute rotation of $v_{p^*}$ is locally optimized by:
\begin{equation} \label{eq:10}
    \bm{R}_{p^*}^*=\arg\min\sum_{\substack{v_n\in\mathcal{V}^s\\e_{n,p^*}\in\mathcal{E}_{p^*}^{m^*}}}d_{\bm{R}}^2(\bm{R}_{n,p^*},\bm{R}_{p^*}\bm{R}_n^{\top}).
\end{equation}
For the global optimization, the inlier edge set is also obtained:
\begin{equation} \label{eq:11}
    \begin{aligned}
        \left(\mathcal{E}_{p^*}\cup\mathcal{E}^s\right)^*=\left\{d_{\bm{R}}
        (\bm{R}_{m,n},\bm{R}_n\bm{R}_m^{\top})<\theta_{\mathrm{th}}\right\}\\
        \mathrm{for}\;v_m,v_n\in\{v_{p^*}\}\cup\mathcal{V}^s,e_{m,n}\in\mathcal{E}_{p^*}\cup\mathcal{E}^s,
    \end{aligned}
\end{equation}
where $\mathcal{E}^s$ is the edge set of $\mathcal{V}^s$. Then, the absolute rotations of the vertices in $\{v_{p^*}\}\cup\mathcal{V}^s$ are globally optimized by:
\begin{equation} \label{eq:12}
    \{\bm{R}_m^*\}=\arg\min\sum_{\substack{v_m,v_n\in\{v_{p^*}\}\cup\mathcal{V}^s\\e_{m,n}\in\left(\mathcal{E}_{p^*}\cup\mathcal{E}^s\right)^*}}d_{\bm{R}}^2(\bm{R}_{m,n},\bm{R}_n\bm{R}_m^{\top}).
\end{equation}

\subsection{Other key steps stemming from the IRAv3+} \label{sec2b}

As the main difference between IRAv3+~\cite{Gao-TCSVT-24} and IRAv4 proposed in this paper is the manner of global reference construction, the other key steps of IRAv4 stemming from IRAv3+ are described in this sub-section for better understanding.

\noindent\textbf{Community detection-based cluster seed construction:} With the objective of performing the following on-the-fly EG clustering and local rotation averaging, several cluster seeds should be constructed in advance for the cluster growing procedure. In both IRAv3~\cite{Gao-TCSVT-23} and IRAv3+, they are constructed based on the community detection method~\cite{Zhou-ECCV-20}. Specifically, the above community detection method is performed on the input EG to generate community-based structure, and for each sub-EG of the community, camera triplet is selected, together with their absolute rotations estimated, which is similar to the initialization step of the task-specific CDS-based global reference construction in IRAv4 (\emph{cf.}~Eq.~\ref{eq:1} to Eq.~\ref{eq:4}).

\noindent\textbf{On-the-fly EG clustering and local rotation averaging:} This step is first adopted in IRAv3, which is similar in principle to the iteration step of the task-specific CDS-based global reference construction in IRAv4 (\emph{cf.}~Eq.~\ref{eq:5} to Eq.~\ref{eq:12}). To achieve this, for each unaffiliated vertex, the edge set between it and the currently affiliated vertex set of each cluster is first obtained. Then, by leveraging these edge sets, the supporting edge set of each unaffiliated vertex within each cluster could be obtained by absolute rotation pre-computation (\emph{cf.}~Eq.~\ref{eq:5}) and relative rotation re-computation (\emph{cf.}~Eq.~\ref{eq:6}). On this basis, the next-best affiliated vertex, together with its affiliated cluster and initialized rotation are simultaneously determined by supporting set global maximization (\emph{cf.}~Eq.~\ref{eq:7} and Eq.~\ref{eq:8}). Then, the rotation estimation are locally (only on the next-best affiliated vertex) or globally (on all the vertices of the affiliated cluster) optimized (\emph{cf.}~Eq.~\ref{eq:9} to Eq.~\ref{eq:12}). The above procedure is iteratively performed until all the vertices are affiliated, together with their rotations estimated.

\noindent\textbf{Common vertices-guided alignment rotation estimation:} After reference construction and EG clustering, cluster-to-reference rotation is estimated for local rotation alignment. As observed in IRAv3+ that the rotation averaging-based absolute rotation estimates tend to be more accurate and reliable than those of the essential matrix decomposition-based relative ones, a common vertices-guided alignment rotation estimation method is proposed. For a particular cluster, either one common vertex or one shared edge between it and the reference could induce an estimate of the cluster-to-reference alignment rotation. As the common vertices-induced ones are with higher priority, they are served as guidance for alignment rotation estimation. Specifically, for each vertex-induced estimate, one could obtain its supporters from the edge-induced ones, and the one with most supporters is used as the initialization of the alignment rotation, which is further optimized by using the constraints provided by its supporters. Readers may refer to the paper of IRAv3+~\cite{Gao-TCSVT-23} for more details.

\begin{table}
\centering
\caption{Metadata of the 1DSfM dataset, where $|\mathcal{V}|$ and $|\mathcal{V}^*|$ indicate the number of vertices in Epipolar-geometry Graph~(EG) and with Ground Truth~(GT), and $|\mathcal{E}|$ indicates the number of edges in EG; $\tilde{r}_{ij}$ and $\bar{r}_{ij}$ denote the median and mean errors in degrees of the relative rotations; and $\tilde{n}_{ij}$ and $\bar{n}_{ij}$ are the median and mean values of the pair-wise feature match number.}
\resizebox{0.6\columnwidth}{!}{
\begin{tabu}{l|c:c:c|c:c|c:c}\tabucline[1pt]{-}
Data & $|\mathcal{V}|$ & $|\mathcal{V}^*|$ & $|\mathcal{E}|$ & $\tilde{r}_{ij}$ & $\bar{r}_{ij}$ & $\tilde{n}_{ij}$ & $\bar{n}_{ij}$\\\tabucline[0.5pt]{-}
ALM & $627$ & $577$ & $97206$ & $2.78\degree$ & $9.09\degree$ & $105$ & $192$\\
ELS & $247$ & $227$ & $20297$ & $2.89\degree$ & $12.50\degree$ & $106$ & $160$\\
GDM & $742$ & $677$ & $48144$ & $12.30\degree$ & $33.33\degree$ & $73$ & $144$\\
MDR & $394$ & $341$ & $23784$ & $9.34\degree$ & $29.30\degree$ & $61$ & $128$\\
MND & $474$ & $450$ & $52424$ & $1.67\degree$ & $7.51\degree$ & $180$ & $310$\\
NYC & $376$ & $332$ & $20680$ & $4.22\degree$ & $14.14\degree$ & $80$ & $167$\\
PDP & $354$ & $338$ & $24710$ & $1.81\degree$ & $8.38\degree$ & $87$ & $128$\\
PIC & $2508$ & $2152$ & $319257$ & $4.93\degree$ & $19.09\degree$ & $56$ & $97$\\
ROF & $1134$ & $1084$ & $70187$ & $2.97\degree$ & $13.83\degree$ & $65$ & $188$\\
TOL & $508$ & $472$ & $23863$ & $2.60\degree$ & $11.58\degree$ & $81$ & $220$\\
TFG & $5433$ & $5058$ & $680012$ & $3.01\degree$ & $8.62\degree$ & $71$ & $109$\\
USQ & $930$ & $789$ & $25561$ & $3.61\degree$ & $9.02\degree$ & $87$ & $150$\\
VNC & $918$ & $836$ & $103550$ & $2.59\degree$ & $11.26\degree$ & $229$ & $408$\\
YKM & $458$ & $437$ & $27729$ & $2.68\degree$ & $11.16\degree$ & $112$ & $245$\\
\tabucline[1pt]{-}
\end{tabu}
}
\label{tab1}
\end{table}

\begin{table*}
\centering
\caption{Real experimental comparison results in absolute rotation estimation on the 1DSfM dataset: \fcolorbox{white}{color_1st}{First}\fcolorbox{white}{color_2nd}{Second}\fcolorbox{white}{color_3rd}{Third}\fcolorbox{white}{color_4th}{Fourth}.}
\resizebox{0.9\textwidth}{!}{
\begin{tabular}{l|c:c:c|c:c:c|c:c:c|c:c:c:c|c}\Xhline{1pt}
\multirow{3}{*}{Data} & \multicolumn{3}{c|}{Robust loss-based} & \multicolumn{3}{c|}{Outlier filtering-based} & \multicolumn{3}{c|}{Deep learning-based} & \multicolumn{4}{c|}{Previous IRA series} & \multirow{3}{*}{IRAv4}\\\Xcline{2-14}{0.5pt}
 & IRLS-$\ell_{\frac{1}{2}}$ & MPLS & {DESC} & OMSTs & HRRA & HARA & NeuRoRA & MSP & RAGO & IRA & IRA++ & IRAv3 & IRAv3+ & \\
 & \cite{Chatterjee-TPAMI-18} & \cite{Shi-ICML-20} & \cite{Shi-ICML-22} & \cite{Cui-ISPRS-19} & \cite{Gao-SPL-20} & \cite{Lee-CVPR-22} & \cite{Purkait-ECCV-20} & \cite{Yang-CVPR-21} & \cite{Li-CVPR-22} & \cite{Gao-IJCV-21} & \cite{Gao-TCSVT-22-1} & \cite{Gao-TCSVT-23} & \cite{Gao-TCSVT-24} & \\\Xhline{0.5pt}
ALM & $2.14\degree$ & $1.16\degree$ & $1.63\degree$ & $1.26\degree$ & $1.03\degree$ & $1.15\degree$ & $1.16\degree$ & $1.07\degree$ & $0.88\degree$ & $0.83\degree$ & \cellcolor{color_4th}$0.80\degree$ & \cellcolor{color_2nd}$0.73\degree$ & \cellcolor{color_1st}$0.72\degree$ & \cellcolor{color_2nd}$0.73\degree$\\
ELS & $1.15\degree$ & $0.88\degree$ & $1.01\degree$ & $0.75\degree$ & $0.59\degree$ & $0.62\degree$ & $0.64\degree$ & $0.83\degree$ & \cellcolor{color_4th}$0.46\degree$ & $0.51\degree$ & \cellcolor{color_4th}$0.46\degree$ & \cellcolor{color_3rd}$0.44\degree$ & \cellcolor{color_1st}$0.39\degree$ & \cellcolor{color_2nd}$0.41\degree$\\
GDM & $28.20\degree$ & $9.87\degree$ & $9.43\degree$ & $45.15\degree$ & $4.04\degree$ & $63.74\degree$ & $2.94\degree$ & $3.69\degree$ & \cellcolor{color_4th}$2.68\degree$ & $5.32\degree$ & $2.88\degree$ & \cellcolor{color_2nd}$1.99\degree$ & \cellcolor{color_3rd}$2.20\degree$ & \cellcolor{color_1st}$1.90\degree$\\      
MDR & $3.08\degree$ & $1.26\degree$ & $2.35\degree$ & $1.12\degree$ & $2.54\degree$ & $1.50\degree$ & $1.13\degree$ & $1.09\degree$ & $1.03\degree$ & $0.85\degree$ & \cellcolor{color_4th}$0.83\degree$ & \cellcolor{color_2nd}$0.75\degree$ & \cellcolor{color_2nd}$0.75\degree$ & \cellcolor{color_1st}$0.74\degree$\\
MND & $0.71\degree$ & $0.51\degree$ & $0.59\degree$ & $0.68\degree$ & $0.62\degree$ & $0.51\degree$ & $0.60\degree$ & $0.50\degree$ & \cellcolor{color_4th}$0.46\degree$ & $0.51\degree$ & $0.50\degree$ & \cellcolor{color_3rd}$0.44\degree$ & \cellcolor{color_1st}$0.40\degree$ & \cellcolor{color_1st}$0.40\degree$\\
NYC & $1.40\degree$ & $1.24\degree$ & $1.48\degree$ & $1.30\degree$ & $1.24\degree$ & $1.37\degree$ & $1.18\degree$ & $1.12\degree$ & \cellcolor{color_1st}$0.71\degree$ & $1.00\degree$ & $0.95\degree$ & \cellcolor{color_4th}$0.82\degree$ & \cellcolor{color_3rd}$0.81\degree$ & \cellcolor{color_2nd}$0.77\degree$\\
PDP & $2.62\degree$ & $1.93\degree$ & $1.95\degree$ & $1.73\degree$ & $0.92\degree$ & $0.92\degree$ & $0.79\degree$ & $0.76\degree$ & \cellcolor{color_2nd}$0.63\degree$ & $0.90\degree$ & $0.75\degree$ & \cellcolor{color_4th}$0.72\degree$ & \cellcolor{color_2nd}$0.63\degree$ & \cellcolor{color_1st}$0.62\degree$\\
PIC & $3.12\degree$ & $1.81\degree$ & $2.43\degree$ & \cellcolor{color_3rd}$1.41\degree$ & $4.87\degree$ & $3.22\degree$ & $1.91\degree$ & $1.80\degree$ & \cellcolor{color_1st}$0.58\degree$ & $1.67\degree$ & $1.70\degree$ & \cellcolor{color_4th}$1.50\degree$ & $1.55\degree$ & $\cellcolor{color_2nd}1.40\degree$\\
ROF & $1.70\degree$ & $1.37\degree$ & $1.42\degree$ & $1.85\degree$ & $2.48\degree$ & $2.42\degree$ & $1.31\degree$ & $1.19\degree$ & \cellcolor{color_3rd}$1.10\degree$ & $1.51\degree$ & $1.24\degree$ & \cellcolor{color_2nd}$1.09\degree$ & \cellcolor{color_4th}$1.14\degree$ & $\cellcolor{color_1st}0.96\degree$\\
TOL & $2.45\degree$ & $2.20\degree$ & $2.49\degree$ & $2.10\degree$ & $2.05\degree$ & $2.36\degree$ & $1.46\degree$ & \cellcolor{color_4th}$1.25\degree$ & \cellcolor{color_3rd}$1.20\degree$ & $2.45\degree$ & $1.33\degree$ & $1.44\degree$ & \cellcolor{color_1st}$1.09\degree$ & \cellcolor{color_2nd}$1.11\degree$\\
TFG & $2.03\degree$ & --- & $1.84\degree$ & $2.63\degree$ & $4.88\degree$ & $2.06\degree$ & $2.25\degree$ & --- & \cellcolor{color_2nd}$1.53\degree$ & $3.30\degree$ & \cellcolor{color_3rd}$1.74\degree$ & \cellcolor{color_1st}$1.49\degree$ & $1.80\degree$ & \cellcolor{color_4th}$1.79\degree$\\
USQ & $4.97\degree$ & $3.48\degree$ & $4.31\degree$ & $3.83\degree$ & $3.77\degree$ & $4.78\degree$ & \cellcolor{color_3rd}$2.01\degree$ & \cellcolor{color_1st}$1.85\degree$ & \cellcolor{color_2nd}$1.92\degree$ & $4.40\degree$ & $3.70\degree$ & $3.27\degree$ & \cellcolor{color_4th}$2.77\degree$ & $3.15\degree$\\
VNC & $4.64\degree$ & $2.83\degree$ & $1.99\degree$ & $3.30\degree$ & $1.84\degree$ & $1.49\degree$ & $1.50\degree$ & $1.10\degree$ & \cellcolor{color_4th}$0.89\degree$ & $1.02\degree$ & $0.94\degree$ & \cellcolor{color_3rd}$0.86\degree$ & \cellcolor{color_1st}$0.76\degree$ & \cellcolor{color_2nd}$0.80\degree$\\
YKM & $1.62\degree$ & $1.45\degree$ & $1.60\degree$ & $1.55\degree$ & $1.57\degree$ & $1.65\degree$ & \cellcolor{color_4th}$0.99\degree$ & \cellcolor{color_1st}$0.91\degree$ & \cellcolor{color_2nd}$0.92\degree$ & $1.57\degree$ & $1.38\degree$ & $1.36\degree$ & \cellcolor{color_3rd}$0.98\degree$ & $1.20\degree$\\
\Xhline{0.5pt}
Rank & $12.71$ & $9.62$ & $11.36$ & $10.29$ & $10.14$ & $10.71$ & $7.71$ & $5.92$ & \cellcolor{color_3rd}$3.14$ & $8.07$ & $5.21$ & \cellcolor{color_4th}$3.43$ & \cellcolor{color_2nd}$2.57$ & \cellcolor{color_1st}$2.21$\\
\Xhline{1pt}
\end{tabular}
}
\label{tab2}
\end{table*}

\noindent\textbf{Absolute rotation global alignment and optimization:} Given the alignment rotation of each cluster, the absolute rotations in the cluster's local coordinate system are globally aligned to the reference's coordinate system. After that, inlier edge set of the original EG is firstly obtained based on the aligned rotations and then used for providing constraints for globally optimizing them (\emph{cf.}~Eq.~\ref{eq:11} and Eq.~\ref{eq:12}).

\section{Experimental evaluation}

In this section, the dataset and metric for evaluation are firstly described, and then the proposed IRAv4 method is evaluated in terms of absolute rotation estimation and alignment reference construction for its effectiveness demonstration.

\subsection{Dataset and metric for evaluation}

The evaluation dataset in this paper is the commonly-used 1DSfM dataset~\cite{Wilson-ECCV-14}, which contains $14$ groups of test data stemmed from the Internet. The metadata of the dataset is listed in Table~\ref{tab1}, where the absolute rotation ground truths of each test data are provided by the SfM results of Bundler~\cite{Snavely-IJCV-08}. The accuracy evaluation metric used in this paper is the same as that in the previous IRA series~\cite{Gao-IJCV-21, Gao-TCSVT-22-1, Gao-TCSVT-23, Gao-TCSVT-24}. Specifically, a best relative rotation between the absolute rotation estimations and ground truths is estimated and used to align these two rotation sets. And then the median alignment error in degree is used as the evaluation metric.

\subsection{Comparison in absolute rotation estimation}

Based on the introduced dataset and metric for evaluation, the performance of the proposed IRAv4 in absolute rotation estimation is validated in the manners of both synthetic and real-world experiments.

\subsubsection{Comparison results on synthetic data}

\begin{figure}[h!]
\centering
\includegraphics[width=0.75\columnwidth]{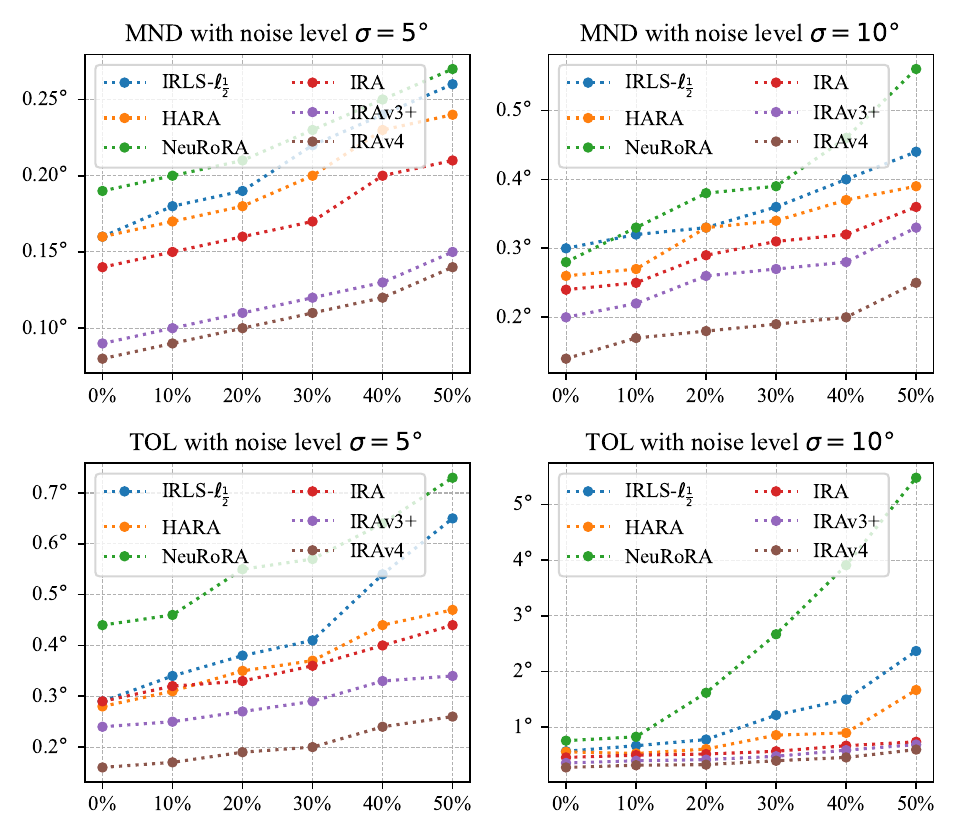}
\caption{Synthetic experimental comparison results in absolute rotation estimation with the graph structures of MND and TOL, where the noise levels of $\sigma$ are set to $5\degree$ and $10\degree$. The $x$-axis and $y$-axis are the rotation median errors and edge outlier percentages.}
\label{fig:2}
\end{figure}

For the synthetic experiments, in order to create evaluation data that as close to the real-world one as possible, similar to CReTA~\cite{Manam-ECCV-22}, the synthetic data is created based on two test data of the 1DSfM dataset, MND and TOL. Specifically, the ground-truth relative rotations of all the epipolar edges are firstly obtained given the ground-truth absolute rotations. Then, following HARA~\cite{Lee-CVPR-22} for each one of the relative rotation, a rotation disturbance whose rotation angle value being in accordance with a normal distribution is added. The standard deviation $\sigma$ is set to either $5\degree$ or $10\degree$. In addition, $p\%$ of the relative rotations are randomly selected and replaced by random rotations to simulate the outlier interruption. $p$ is set to $0$, $10$, $20$, $30$, $40$, and $50$. For comparison, 1 representative method for all 3 kinds of rotation averaging (robust loss-based, outlier filtering-based, and deep learning-based) methods is selected, including IRLS-$\ell_{\frac{1}{2}}$~\cite{Chatterjee-TPAMI-18}, HARA~\cite{Lee-CVPR-22}, and NeuRoRA~\cite{Purkait-ECCV-20}. In addition, the primary method and latest one of the IRA series, \emph{i.e.} IRA~\cite{Gao-IJCV-21} and IRAv3+~\cite{Gao-TCSVT-24}, are also involved in the comparison. The synthetic experimental comparison results between our proposed IRAv4 and others are shown in Fig.~\ref{fig:2}.

It could be observed from the table that: On the one hand, under the same synthetic experiment configurations (noise level and outlier ratio), the rotation estimation accuracy on MND is higher than that of TOL on the whole; On the other hand, the accuracy drops when either the noise level or the outlier ratio increases. It should be further noted that the number of cameras is similar for MND and TOL ($474$ \emph{vs.} $508$), while the number of edges differs obviously ($52424$ \emph{vs.} $23863$). As a result, the hardness of the rotation averaging problem is highly related to noise level and graph density~\cite{Wilson-ECCV-16}. In addition, under all experiment configurations, our IRAv4 achieves best performance, which demonstrates its effectiveness in estimation accuracy and robustness to both estimation noise and measurement outliers.

\subsubsection{Comparison results on real-world data}

For the real-world comparative experiments, except for the previous IRA series, including IRA~\cite{Gao-IJCV-21}, IRA++~\cite{Gao-TCSVT-22-1}, IRAv3~\cite{Gao-TCSVT-23}, and IRAv3+~\cite{Gao-TCSVT-24} which have been described in Sec.~\ref{sec:2}, IRAv4 is also compared with some other mainstream rotation averaging methods, including 3 robust loss-based methods, IRLS-$\ell_{\frac{1}{2}}$~\cite{Chatterjee-TPAMI-18}, MPLS~\cite{Shi-ICML-20}, {DESC~\cite{Shi-ICML-22}}, 3 outlier filtering-based methods, OMSTs~\cite{Cui-ISPRS-19}, HRRA~\cite{Gao-SPL-20}, and HARA~\cite{Lee-CVPR-22}, and 3 deep learning-based methods, NeuRoRA~\cite{Purkait-ECCV-20}, MSP~\cite{Yang-CVPR-21}, and RAGO~\cite{Li-CVPR-22}. The comparative results are illustrated in Table~\ref{tab2}, and to deliver a comprehensive comparison, the rankings of the comparative methods on each test data are averaged and shown in the last row of the table. It could be observed that IRAv4 achieves overall best performance among all the comparative methods, which demonstrates its advantages in absolute rotation estimation.

\begin{table}
\centering
\caption{Experimental comparison results in absolute location estimation on the 1DSfM dataset: \fcolorbox{white}{color_1st}{First}\fcolorbox{white}{color_2nd}{Second}\fcolorbox{white}{color_3rd}{Third}\fcolorbox{white}{color_4th}{Fourth}.}
\resizebox{0.6\columnwidth}{!}{
\begin{tabular}{l|c:c:c|c:c|c}\Xhline{1pt}
\multirow{2}{*}{Data} & IRLS-$\ell_{\frac{1}{2}}$ & HARA & NeuRoRA & IRA & IRAv3+ & \multirow{2}{*}{IRAv4}\\
 & \cite{Chatterjee-TPAMI-18} & \cite{Lee-CVPR-22} & \cite{Purkait-ECCV-20} & \cite{Gao-IJCV-21} & \cite{Gao-TCSVT-24} & \\\Xhline{0.5pt}
ALM & $1.11\mathrm{m}$ & \cellcolor{color_4th}$0.84\mathrm{m}$ & $0.85\mathrm{m}$ & \cellcolor{color_3rd}$0.70\mathrm{m}$ & \cellcolor{color_1st}$0.66\mathrm{m}$ & \cellcolor{color_2nd}$0.67\mathrm{m}$\\
ELS & $6.55\mathrm{m}$ & \cellcolor{color_4th}$4.89\mathrm{m}$ & $4.96\mathrm{m}$ & \cellcolor{color_3rd}$4.49\mathrm{m}$ & \cellcolor{color_2nd}$3.96\mathrm{m}$ & \cellcolor{color_1st}$3.91\mathrm{m}$\\
GDM & $7.82\mathrm{m}$ & $15.41\mathrm{m}$ & \cellcolor{color_4th}$6.29\mathrm{m}$ & \cellcolor{color_3rd}$2.78\mathrm{m}$ & \cellcolor{color_2nd}$2.62\mathrm{m}$ & \cellcolor{color_1st}$2.42\mathrm{m}$\\
MDR & $7.67\mathrm{m}$ & $5.20\mathrm{m}$ & \cellcolor{color_4th}$4.86\mathrm{m}$ & \cellcolor{color_2nd}$3.86\mathrm{m}$ & \cellcolor{color_3rd}$4.13\mathrm{m}$ & \cellcolor{color_1st}$3.66\mathrm{m}$\\
MND & $0.92\mathrm{m}$ & \cellcolor{color_3rd}$0.79\mathrm{m}$ & $0.88\mathrm{m}$ & \cellcolor{color_3rd}$0.79\mathrm{m}$ & \cellcolor{color_2nd}$0.71\mathrm{m}$ & \cellcolor{color_1st}$0.70\mathrm{m}$\\
NYC & $1.43\mathrm{m}$ & \cellcolor{color_4th}$1.37\mathrm{m}$ & $1.53\mathrm{m}$ & \cellcolor{color_3rd}$1.26\mathrm{m}$ & \cellcolor{color_2nd}$1.19\mathrm{m}$ & \cellcolor{color_1st}$1.18\mathrm{m}$\\
PDP & $1.73\mathrm{m}$ & \cellcolor{color_4th}$1.05\mathrm{m}$ & \cellcolor{color_3rd}$0.98\mathrm{m}$ & $1.06\mathrm{m}$ & \cellcolor{color_2nd}$0.91\mathrm{m}$ & \cellcolor{color_1st}$0.88\mathrm{m}$\\
PIC & \cellcolor{color_4th}$2.42\mathrm{m}$ & $2.49\mathrm{m}$ & $4.05\mathrm{m}$ & \cellcolor{color_3rd}$1.77\mathrm{m}$ & \cellcolor{color_2nd}$1.72\mathrm{m}$ & \cellcolor{color_1st}$1.67\mathrm{m}$\\
ROF & \cellcolor{color_4th}$7.09\mathrm{m}$ & $8.40\mathrm{m}$ & $9.74\mathrm{m}$ & \cellcolor{color_3rd}$6.28\mathrm{m}$ & \cellcolor{color_2nd}$6.15\mathrm{m}$ & \cellcolor{color_1st}$5.47\mathrm{m}$\\
TOL & $4.16\mathrm{m}$ & $4.11\mathrm{m}$ & \cellcolor{color_3rd}$3.10\mathrm{m}$ & \cellcolor{color_4th}$3.86\mathrm{m}$ & \cellcolor{color_1st}$2.59\mathrm{m}$ & \cellcolor{color_2nd}$2.76\mathrm{m}$\\
TFG & \cellcolor{color_2nd}$6.77\mathrm{m}$ & \cellcolor{color_3rd}$6.80\mathrm{m}$ & $12.21\mathrm{m}$ & \cellcolor{color_4th}$8.66\mathrm{m}$ & $9.37\mathrm{m}$ & \cellcolor{color_1st}$6.37\mathrm{m}$\\
USQ & $8.06\mathrm{m}$ & \cellcolor{color_4th}$7.87\mathrm{m}$ & $11.86\mathrm{m}$ & \cellcolor{color_3rd}$7.36\mathrm{m}$ & \cellcolor{color_1st}$5.89\mathrm{m}$ & \cellcolor{color_2nd}$6.55\mathrm{m}$\\
VNC & $6.17\mathrm{m}$ & \cellcolor{color_4th}$3.36\mathrm{m}$ & $3.90\mathrm{m}$ & \cellcolor{color_3rd}$2.81\mathrm{m}$ & \cellcolor{color_1st}$2.44\mathrm{m}$ & \cellcolor{color_2nd}$2.76\mathrm{m}$\\
YKM & \cellcolor{color_4th}$2.04\mathrm{m}$ & $2.11\mathrm{m}$ & \cellcolor{color_2nd}$1.74\mathrm{m}$ & $2.05\mathrm{m}$ & \cellcolor{color_1st}$1.63\mathrm{m}$ & \cellcolor{color_3rd}$1.84\mathrm{m}$\\
\Xhline{0.5pt}
Rank & $5.07$ & \cellcolor{color_4th}$4.43$ & $4.71$ & \cellcolor{color_3rd}$3.36$ & \cellcolor{color_2nd}$1.93$ & \cellcolor{color_1st}$1.43$\\
\Xhline{1pt}
\end{tabular}
}
\label{tab3}
\end{table}

\begin{figure}
\centering
\includegraphics[width=0.75\columnwidth]{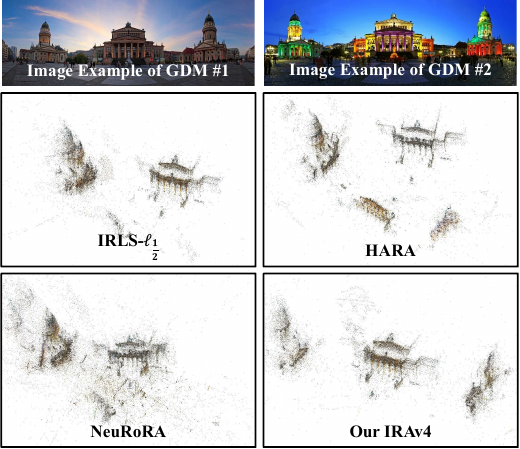}
\caption{Sparse scene reconstruction results on GDM by $4$ comparative rotation averaging methods, including IRLS-$\ell_{\frac{1}{2}}$~\cite{Chatterjee-TPAMI-18}, HARA~\cite{Lee-CVPR-22}, NeuRoRA~\cite{Purkait-ECCV-20}, and IRAv4, combined with translation averaging (BATA~\cite{Zhuang-CVPR-18}), multi-view triangulation~\cite{Schonberger-CVPR-16}, and global bundle adjustment~\cite{Agarwal-ECCV-10}.}
\label{fig:3}
\end{figure}

In addition, to give a more comprehensive comparison on different rotation averaging methods, the absolute rotations obtained by them are fed into a representative translation averaging method for absolute camera location recovery. The methods for comparison are the same as those in the synthetic experiments, and the results are shown in Table~\ref{tab3}, where the median alignment errors in camera locations wrt. the results of Bundler~\cite{Snavely-IJCV-08} are listed. It could be observed that by leveraging the absolute rotations provided by IRAv4, camera locations those with highest accuracy is achieved, especially for the most difficult test data, GDM, where symmetrical scene structure and numerous relative motion measurement outliers exist. Given the rotation and location estimates of GDM, multi-view triangulation~\cite{Schonberger-CVPR-16} and global bundle adjustment~\cite{Agarwal-ECCV-10} are performed for camera pose and scene structure optimization. And the sparse reconstruction results of GDM by IRLS-$\ell_{\frac{1}{2}}$, HARA, NeuRoRA, and IRAv4-based global SfM are shown in Fig.~\ref{fig:3}. It could be observed that among the $4$ global SfM variants, only the one based on our IRAv4 could successfully recover the correct scene structure of GDM, which further validates its robustness to scenes with ambiguous structures and EG with high noise.

\begin{table}
\scriptsize
\centering
\caption{Comparison in alignment reference construction on the 1DSfM dataset, $n_{\mathrm{ref}}$ and $e_{\mathrm{ref}}$ are reference's size and accuracy.}
\resizebox{0.7\columnwidth}{!}{
\begin{tabular}{l|c:c|c:c|c:c}\Xhline{1pt}
\multirow{2}{*}{Data} & \multicolumn{2}{c|}{WCDS~\cite{Jiang-TGRS-22}} & \multicolumn{2}{c|}{IRAv3+~\cite{Gao-TCSVT-24}} & \multicolumn{2}{c}{IRAv4}\\\Xcline{2-7}{0.5pt}
& $n_{\mathrm{ref}}$ & $e_{\mathrm{ref}}$ & $n_{\mathrm{ref}}$ & $e_{\mathrm{ref}}$ & $n_{\mathrm{ref}}$ & $e_{\mathrm{ref}}$\\
\Xhline{0.5pt}
ALM & $47$ & $2.01\degree$ & $95.2\pm1.8$ & $1.16\degree\pm0.14\degree$ & $72$ & $0.92\degree$\\
ELS & $16$ & $0.89\degree$ & $34.9\pm1.3$ & $1.20\degree\pm0.11\degree$ & $20$ & $0.63\degree$\\
GDM & $65$ & $4.56\degree$ & $120.7\pm6.3$ & $2.70\degree\pm1.06\degree$ & $105$ & $2.38\degree$\\
MDR & $44$ & $1.05\degree$ & $61.7\pm2.5$ & $0.83\degree\pm0.26\degree$ & $63$ & $0.97\degree$\\
MND & $32$ & $1.95\degree$ & $65.7\pm2.8$ & $0.98\degree\pm0.35\degree$ & $57$ & $0.61\degree$\\
NYC & $42$ & $3.93\degree$ & $58.2\pm2.5$ & $0.98\degree\pm0.11\degree$ & $56$ & $0.55\degree$\\
PDP & $32$ & $2.04\degree$ & $55.3\pm1.6$ & $1.46\degree\pm0.22\degree$ & $52$ & $1.15\degree$\\
PIC & $208$ & $2.12\degree$ & $399.0\pm4.6$ & $2.65\degree\pm0.38\degree$ & $363$ & $1.55\degree$\\
ROF & $96$ & $4.02\degree$ & $175.8\pm3.3$ & $1.58\degree\pm0.38\degree$ & $181$ & $0.80\degree$\\
TOL & $57$ & $3.39\degree$ & $78.6\pm2.2$ & $1.00\degree\pm0.12\degree$ & $71$ & $0.98\degree$\\
TFG & $696$ & $4.08\degree$ & $886.1\pm7.4$ & $2.12\degree\pm0.40\degree$ & $953$ & $2.86\degree$\\
USQ & $136$ & $3.49\degree$ & $171.0\pm3.4$ & $3.16\degree\pm0.50\degree$ & $181$ & $3.23\degree$\\
VNC & $76$ & $4.55\degree$ & $144.1\pm3.5$ & $1.96\degree\pm0.57\degree$ & $135$ & $1.38\degree$\\
YKM & $42$ & $1.23\degree$ & $70.4\pm2.2$ & $1.55\degree\pm0.19\degree$ & $64$ & $1.22\degree$\\
\Xhline{1pt}
\end{tabular}
}
\label{tab4}
\end{table}

\subsection{Comparison in alignment reference construction}

As the most critical technique in IRAv4, the task-specific alignment reference construction method is compared with other two used in WCDS~\cite{Jiang-TGRS-22} and IRAv3+~\cite{Gao-TCSVT-24}, in terms of reference size and accuracy. The comparison results are shown in Table~\ref{tab3}. It should be noted that for the method in IRAv3+, multiple CDSs are randomly extracted, by which different alignment references are constructed for different trails. Therefore, for the result of IRAv3+ in the table, $10$ trails reference construction by IRAv3+ are conducted and the mean value and standard deviation are listed. From the table we could see that: 1)~On the one hand, compared with the naive method employed in WCDS~\cite{Jiang-TGRS-22}, though reference with larger size is extracted by IRAv4, its accuracy is much higher; 2)~On the other hand, compared with the method proposed in IRAv3+, though with similar values in both $n_{\mathrm{ref}}$ and $e_{\mathrm{ref}}$, the reference constructed by IRAv4 is with smaller size and higher accuracy on the whole. As a result, the effectiveness of the proposed reference construction method is validated.

\section{Conclusion}

In this paper, to benefit from the global alignment reference for local-to-global rotation alignment more, an incremental parameter estimation-based task-specific connected dominating set extraction method is proposed, by which the accuracy and robustness of the Incremental Rotation Averaging~(IRA) series are further advanced. The proposed method in this paper, IRAv4, is comprehensively evaluated on the 1DSfM dataset and achieves overall the best performance in both absolute rotation estimation and alignment reference construction, which demonstrates its effectiveness when dealing with large-scale and high-noise rotation averaging problems.

\section*{CRediT authorship contribution statement}

\textbf{Xiang Gao:} Conceptualization, Formal analysis, Methodology, Software, Validation, Writing – original draft. \textbf{Hainan Cui:} Data curation, Investigation. \textbf{Yangdong Liu:} Project administration, Visualization. \textbf{Shuhan Shen:} Funding acquisition, Resources, Supervision, Writing – review \& editing.

\section*{Acknowledgements}

This work was supported by the National Science Foundation of China under Grant 62373349, Grant U22B2055, and Grant 62402495, and the Key R\&D Project in Henan Province under Grant 231111210300.

\section*{Declaration of competing interest}

The authors have no competing interests to declare that are relevant to the content of this article.

\section*{Data availability}

Data will be made available on request.

{\small
\bibliographystyle{elsarticle-num}
\bibliography{xgao-lib}

\begin{thebibliography}{10}
\expandafter\ifx\csname url\endcsname\relax
  \def\url#1{\texttt{#1}}\fi
\expandafter\ifx\csname urlprefix\endcsname\relax\def\urlprefix{URL }\fi
\expandafter\ifx\csname href\endcsname\relax
  \def\href#1#2{#2} \def\path#1{#1}\fi

\bibitem{Gao-ISPRS-18}
X.~Gao, S.~Shen, Y.~Zhou, H.~Cui, L.~Zhu, Z.~Hu, Ancient {Chinese} architecture {3D} preservation by merging ground and aerial point clouds, ISPRS Journal of Photogrammetry and Remote Sensing 143 (2018) 72--84.

\bibitem{Gao-TCSVT-20}
X.~Gao, S.~Shen, L.~Zhu, T.~Shi, Z.~Wang, Z.~Hu, Complete scene reconstruction by merging images and laser scans, IEEE Transactions on Circuits and Systems for Video Technology 30~(10) (2020) 3688--3701.

\bibitem{Gao-TCSVT-22-2}
X.~Gao, L.~Zhu, B.~Fan, H.~Liu, S.~Shen, Incremental translation averaging, IEEE Transactions on Circuits and Systems for Video Technology 32~(11) (2022) 7783--7795.

\bibitem{Gao-TITS-23}
X.~Gao, D.~Tao, Y.~Liu, Z.~Xie, S.~Shen, Vehicle-borne multi-sensor temporal-spatial pose globalization via cross-domain data association, IEEE Transactions on Intelligent Transportation Systems 24~(12) (2023) 13962--13975.

\bibitem{Cui-ICCV-15}
Z.~Cui, P.~Tan, Global structure-from-motion by similarity averaging, in: IEEE International Conference on Computer Vision (ICCV), 2015, pp. 864--872.

\bibitem{Schonberger-CVPR-16}
J.~L. Schönberger, J.~M. Frahm, Structure-from-motion revisited, in: IEEE Conference on Computer Vision and Pattern Recognition (CVPR), 2016, pp. 4104--4113.

\bibitem{Cui-CVPR-17}
H.~Cui, X.~Gao, S.~Shen, Z.~Hu, {HSfM}: Hybrid structure-from-motion, in: IEEE Conference on Computer Vision and Pattern Recognition (CVPR), 2017, pp. 2393--2402.

\bibitem{Cui-3DV-17}
H.~Cui, S.~Shen, X.~Gao, Z.~Hu, Batched incremental structure-from-motion, in: International Conference on 3D Vision (3DV), 2017, pp. 205--214.

\bibitem{Zhu-CVPR-18}
S.~{Zhu}, R.~{Zhang}, L.~{Zhou}, T.~{Shen}, T.~{Fang}, P.~{Tan}, L.~{Quan}, Very large-scale global {SfM} by distributed motion averaging, in: IEEE Conference on Computer Vision and Pattern Recognition (CVPR), 2018, pp. 4568--4577.

\bibitem{Manam-ECCV-22}
L.~Manam, V.~M. Govindu, Correspondence reweighted translation averaging, in: European Conference on Computer Vision (ECCV), 2022, pp. 56--72.

\bibitem{Cui-TIP-23}
H.~Cui, X.~Gao, S.~Shen, {MCSfM}: Multi-camera-based incremental structure-from-motion, IEEE Transactions on Image Processing 32 (2023) 6441--6456.

\bibitem{Nister-TPAMI-04}
D.~{Nister}, An efficient solution to the five-point relative pose problem, IEEE Transactions on Pattern Analysis and Machine Intelligence 26~(6) (2004) 756--770.

\bibitem{Govindu-CVPR-01}
V.~M. {Govindu}, Combining two-view constraints for motion estimation, in: IEEE Conference on Computer Vision and Pattern Recognition (CVPR), 2001, pp. 218--225.

\bibitem{Martinec-CVPR-07}
D.~Martinec, T.~Pajdla, Robust rotation and translation estimation in multiview reconstruction, in: IEEE Conference on Computer Vision and Pattern Recognition (CVPR), 2007, pp. 1--8.

\bibitem{Crandall-TPAMI-13}
D.~{Crandall}, A.~{Owens}, N.~{Snavely}, D.~{Huttenlocher}, {SfM with MRFs: Discrete}-continuous optimization for large-scale structure from motion, IEEE Transactions on Pattern Analysis and Machine Intelligence 35~(12) (2013) 2841--2853.

\bibitem{Moulon-ICCV-13}
P.~{Moulon}, P.~{Monasse}, R.~{Marlet}, Global fusion of relative motions for robust, accurate and scalable structure from motion, in: IEEE International Conference on Computer Vision (ICCV), 2013, pp. 3248--3255.

\bibitem{Pan-ECCV-24-1}
L.~Pan, D.~Barath, M.~Pollefeys, J.~L. Schönberger, Global structure-from-motion revisited, in: European Conference on Computer Vision (ECCV), 2024, pp. 1--19.

\bibitem{Hartley-IJCV-13}
R.~Hartley, J.~Trumpf, Y.~Dai, H.~Li, Rotation averaging, International Journal of Computer Vision 103 (2013) 267–305.

\bibitem{Ozyesil-CVPR-15}
O.~{Özyeşil}, A.~{Singer}, Robust camera location estimation by convex programming, in: IEEE Conference on Computer Vision and Pattern Recognition (CVPR), 2015, pp. 2674--2683.

\bibitem{Wilson-CVPR-20}
K.~Wilson, D.~Bindel, On the distribution of minima in intrinsic-metric rotation averaging, in: IEEE/CVF Conference on Computer Vision and Pattern Recognition (CVPR), 2020, pp. 6030--6038.

\bibitem{Dellaert-ECCV-20}
F.~Dellaert, D.~M. Rosen, J.~Wu, R.~Mahony, L.~Carlone, Shonan rotation averaging: Global optimality by surfing ${SO}(p)^n$, in: European Conference on Computer Vision (ECCV), 2020, pp. 292--308.

\bibitem{Eriksson-TPAMI-21}
A.~{Eriksson}, C.~{Olsson}, F.~{Kahl}, T.~J. {Chin}, Rotation averaging with the chordal distance: Global minimizers and strong duality, IEEE Transactions on Pattern Analysis and Machine Intelligence 43~(1) (2021) 256--268.

\bibitem{Chen-CVPR-21}
Y.~Chen, J.~Zhao, L.~Kneip, Hybrid rotation averaging: {A} fast and robust rotation averaging approach, in: IEEE/CVF Conference on Computer Vision and Pattern Recognition (CVPR), 2021, pp. 10358--10367.

\bibitem{Parra-CVPR-21}
A.~Parra, S.-F. Chng, T.-J. Chin, A.~Eriksson, I.~Reid, Rotation coordinate descent for fast globally optimal rotation averaging, in: IEEE/CVF Conference on Computer Vision and Pattern Recognition (CVPR), 2021, pp. 4296--4305.

\bibitem{Moreira-ICCV-21}
G.~Moreira, M.~Marques, J.~P. Costeira, Rotation averaging in a split second: A primal-dual method and a closed-form for cycle graphs, in: IEEE/CVF International Conference on Computer Vision (ICCV), 2021, pp. 5432--5440.

\bibitem{Zhang-CVPR-23}
G.~Zhang, V.~Larsson, D.~Barath, Revisiting rotation averaging: Uncertainties and robust losses, in: IEEE/CVF Conference on Computer Vision and Pattern Recognition (CVPR), 2023, pp. 17215--17224.

\bibitem{Li-CVPR-24}
S.~Li, Y.~Shi, G.~Lerman, Efficient detection of long consistent cycles and its application to distributed synchronization, in: IEEE/CVF Conference on Computer Vision and Pattern Recognition (CVPR), 2024, pp. 5260--5269.

\bibitem{Pan-ECCV-24-2}
L.~Pan, M.~Pollefeys, D.~Barath, Gravity-aligned rotation averaging with circular regression, in: European Conference on Computer Vision (ECCV), 2024, pp. 1--19.

\bibitem{Gao-IJCV-21}
X.~Gao, L.~Zhu, Z.~Xie, H.~Liu, S.~Shen, Incremental rotation averaging, International Journal of Computer Vision 129 (2021) 1202--1216.

\bibitem{Gao-TCSVT-22-1}
X.~Gao, L.~Zhu, H.~Cui, Z.~Xie, S.~Shen, {IRA++: Distributed} incremental rotation averaging, IEEE Transactions on Circuits and Systems for Video Technology 32~(7) (2022) 4885--4892.

\bibitem{Gao-TCSVT-23}
X.~Gao, H.~Cui, M.~Li, Z.~Xie, S.~Shen, {IRAv3}: Hierarchical incremental rotation averaging on the fly, IEEE Transactions on Circuits and Systems for Video Technology 33~(4) (2023) 2001--2006.

\bibitem{Gao-TCSVT-24}
X.~Gao, H.~Cui, W.~Huang, M.~Li, S.~Shen, {IRAv3+}: Hierarchical incremental rotation averaging via multiple connected dominating sets, IEEE Transactions on Circuits and Systems for Video Technology 34~(4) (2024) 3049--3055.

\bibitem{Jiang-TGRS-22}
S.~Jiang, Q.~Li, W.~Jiang, W.~Chen, Parallel structure from motion for {UAV} images via weighted connected dominating set, IEEE Transactions on Geoscience and Remote Sensing 60 (2022) 5413013.

\bibitem{Chatterjee-TPAMI-18}
A.~Chatterjee, V.~M. Govindu, Robust relative rotation averaging, IEEE Transactions on Pattern Analysis and Machine Intelligence 40~(4) (2018) 958--972.

\bibitem{Shi-ICML-20}
Y.~Shi, G.~Lerman, Message passing least squares framework and its application to rotation synchronization, in: International Conference on Machine Learning (ICML), 2020, pp. 8796--8806.

\bibitem{Shi-ICML-22}
Y.~Shi, C.~Wyeth, G.~Lerman, Robust group synchronization via quadratic programming, in: International Conference on Machine Learning (ICML), 2022, pp. 20095--20105.

\bibitem{Cui-ISPRS-19}
H.~Cui, S.~Shen, W.~Gao, H.~Liu, Z.~Wang, Efficient and robust large-scale structure-from-motion via track selection and camera prioritization, ISPRS Journal of Photogrammetry and Remote Sensing 156 (2019) 202--214.

\bibitem{Gao-SPL-20}
X.~{Gao}, J.~{Luo}, K.~{Li}, Z.~{Xie}, Hierarchical {RANSAC}-based rotation averaging, IEEE Signal Processing Letters 27 (2020) 1874--1878.

\bibitem{Lee-CVPR-22}
S.~H. Lee, J.~Civera, {HARA}: A hierarchical approach for robust rotation averaging, in: IEEE/CVF Conference on Computer Vision and Pattern Recognition (CVPR), 2022, pp. 15777--15786.

\bibitem{Purkait-ECCV-20}
P.~Purkait, T.~J. Chin, I.~Reid, {NeuRoRA}: Neural robust rotation averaging, in: European Conference on Computer Vision (ECCV), 2020, pp. 137--154.

\bibitem{Yang-CVPR-21}
L.~Yang, H.~Li, J.~A. Rahim, Z.~Cui, P.~Tan, End-to-end rotation averaging with multi-source propagation, in: IEEE/CVF Conference on Computer Vision and Pattern Recognition (CVPR), 2021, pp. 11774--11783.

\bibitem{Li-CVPR-22}
H.~Li, Z.~Cui, S.~Liu, P.~Tan, {RAGO}: Recurrent graph optimizer for multiple rotation averaging, in: IEEE/CVF Conference on Computer Vision and Pattern Recognition (CVPR), 2022, pp. 15787--15796.

\bibitem{Zhou-ECCV-20}
L.~{Zhou}, Z.~{Luo}, M.~{Zhen}, T.~{Shen}, S.~{Li}, Z.~{Huang}, T.~{Fang}, L.~{Quan}, Stochastic bundle adjustment for efficient and scalable {3D} reconstruction, in: European Conference on Computer Vision (ECCV), 2020, pp. 364--379.

\bibitem{Guha-Algorithmica-98}
S.~Guha, S.~Khuller, Approximation algorithms for connected dominating sets, Algorithmica 20 (1998) 374–387.

\bibitem{Wilson-ECCV-14}
K.~Wilson, N.~Snavely, Robust global translations with {1DSfM}, in: European Conference on Computer Vision (ECCV), 2014, pp. 61--75.

\bibitem{Snavely-IJCV-08}
N.~Snavely, S.~M. Seitz, R.~Szeliski, Modeling the world from {Internet} photo collections, International Journal of Computer Vision 80~(2) (2008) 189--210.

\bibitem{Wilson-ECCV-16}
K.~Wilson, D.~Bindel, N.~Snavely, When is rotations averaging hard?, in: European Conference on Computer Vision (ECCV), 2016, pp. 255--270.

\bibitem{Zhuang-CVPR-18}
B.~{Zhuang}, L.~{Cheong}, G.~H. {Lee}, Baseline desensitizing in translation averaging, in: IEEE/CVF Conference on Computer Vision and Pattern Recognition (CVPR), 2018, pp. 4539--4547.

\bibitem{Agarwal-ECCV-10}
S.~Agarwal, N.~Snavely, S.~M. Seitz, R.~Szeliski, Bundle adjustment in the large, in: European Conference on Computer Vision (ECCV), 2010, pp. 29--42.

\end{thebibliography}
}

\end{document}